\documentclass[sigconf]{acmart}

\AtBeginDocument{%
  \providecommand\BibTeX{{%
    \normalfont B\kern-0.5em{\scshape i\kern-0.25em b}\kern-0.8em\TeX}}}

\usepackage{algorithmic}
\usepackage{algorithm}
\usepackage{tabularx}
\usepackage{multirow}
\usepackage{lineno}

\begin{document}


\title{Struggle with Adversarial Defense? Try Diffusion}


\author{Anonymous Authors}


\begin{abstract}
  Adversarial attacks induce misclassification by introducing subtle perturbations. Recently, diffusion models are applied to the image classifiers to improve adversarial robustness through adversarial training or by purifying adversarial noise. However, diffusion-based adversarial training often encounters convergence challenges and high computational expenses. Additionally, diffusion-based purification inevitably causes data shift and is deemed susceptible to stronger adaptive attacks.

To tackle these issues, we propose the Truth Maximization Diffusion Classifier (TMDC), a generative Bayesian classifier that builds upon pre-trained diffusion models and the Bayesian theorem. Unlike data-driven classifiers, TMDC, guided by Bayesian principles, utilizes the conditional likelihood from diffusion models to determine the class probabilities of input images, thereby insulating against the influences of data shift and the limitations of adversarial training. Moreover, to enhance TMDC's resilience against more potent adversarial attacks, we propose an optimization strategy for diffusion classifiers. This strategy involves post-training the diffusion model on perturbed datasets with ground-truth labels as conditions, guiding the diffusion model to learn the data distribution and maximizing the likelihood under the ground-truth labels. The proposed method achieves state-of-the-art performance on the CIFAR10 dataset against heavy white-box attacks and strong adaptive attacks. Specifically, TMDC achieves robust accuracies of 82.81\% against $l_{\infty}$ norm-bounded perturbations and 86.05\% against $l_{2}$ norm-bounded perturbations, respectively, with $\epsilon=0.05$. 
\end{abstract}

\ccsdesc[500]{Computing methodologies~Computer vision problems}
\ccsdesc[500]{Security and privacy~Software security engineering}

\ccsdesc[500]{Security and privacy~Software security engineering}
\ccsdesc[500]{Computing methodologies~Computer vision problems}

\keywords{Diffusion Model, Adversarial Robustness, Generative Classifier}



\maketitle

\section{Introduction}
Since the inception of ImageNet \cite{deng2009imagenet} and its associated competitions, researchers have made significant strides in image classification tasks, particularly with deep neural networks achieving notable success in this domain. Previous endeavors have consistently deepened and broadened networks \cite{krizhevsky2012imagenet,lecun1989handwritten,szegedy2015going,zagoruyko2016wide}, employed residual structures \cite{he2016deep, zagoruyko2016wide}, and utilized transformer architectures \cite{vaswani2017attention,liu2021swin,dosovitskiy2020image}. These progressively refined models consistently establish new benchmarks across significant datasets, showcasing exceptional performance. However, these models are trained and evaluated on samples from natural datasets, rendering them susceptible to disruptions. Adversarial attacks adeptly introduce imperceptible perturbations into image data, leading to misclassification by neural networks and yielding wholly inaccurate outcomes. Consequently, adversarial attacks have emerged as a common evaluation method for assessing model robustness.

Given the crucial role of image classification tasks in fields such as facial recognition \cite{kim2022adaface, an2022killing}, medical health \cite{hashimoto2020multi,maksoud2020sos}, and remote sensing \cite{xu2023deep,kuckreja2023geochat}, the defense against adversarial attacks emerges as a key security concern. Presently, common defensive strategies include adversarial training and image denoising. Notably, the purification approach, which employs diffusion models for denoising, has exhibited promising outcomes. This technique entails utilizing a diffusion model for the generation of image samples through noise addition and subsequent denoising processes, intended for classification or adversarial training purposes. However, it is susceptible to high-intensity adaptive attacks, and the classification performance of the classifier on images post-purification remains suboptimal. We contend that a limiting factor constraining further augmentation of diffusion-based purification efficacy lies in the necessity for images processed by the diffusion model to undergo subsequent inference by discriminative classifiers, that is, in other words, the efficacy of the purification method is partly constrained by the classifier. The noise addition and denoising processes of the diffusion model may disrupt the data distribution of original images, which adheres to the data boundaries learned by the classifier, thereby impeding performance enhancement. Hence, one might inquire, \textbf{why not utilize the diffusion model alone directly for image classification?}

Diffusion models represent a contemporary class of powerful image generation models, distinguished by their inference process comprising forward diffusion and backward denoising stages predominantly. In the forward process, the model systematically introduces Gaussian noise to the image, whereas in the backward process, it undertakes denoising of the perturbed data. Throughout the training phase, Gaussian noise parameters are parameterized utilizing Evidence Lower Bound (ELBO) \cite{blei2017variational}. The diffusion model utilizes neural networks to predict the Gaussian noise added during the forward process to the samples and compute the loss against the ground truth. Previous research has transformed the Stable Diffusion, a conditional diffusion model, into a generative classifier known as the Diffusion Classifier \cite{li2023your}, leveraging Bayesian theorem and computing Monte Carlo estimates for the noise predictions of each class. Li et al. \cite{li2023your} scrutinized its zero-shot performance as a classifier, whereas our study, differently, delves into the adversarial robustness of the Diffusion Classifier.

During the inference process of the Diffusion Classifier, each class label undergoes transformation into prompts that are fed into the model, directing it to infer parameterized noise predictions and compute losses against the ground truth. Subsequently, unbiased Monte Carlo estimates of the expected losses for each class are derived, and the final classification outcome is obtained through Bayesian theorem. Conceptually, this inference process entails comparing the relative magnitudes of model inference losses under different prompts. Hence, theoretically, it can be posited that adversarial attacks, which involve perturbations constrained by norms added to original images, would not significantly impact the inference outcomes of the Diffusion Classifier. Consequently, we propose the assertion that the Diffusion Classifier exhibits adversarial robustness, a proposition substantiated by empirical evidence. Furthermore, we introduce the \textbf{Truth Maximization} optimization method. 
This approach involves training the model with adversarially perturbed input data and conditioning it on text prompts composed of ground-truth labels. The objective is to minimize the prediction loss of parameterized noise in the diffusion process, thereby optimizing model parameters, which enables the model to learn the ability to accurately model image data into the correct categories under adversarial perturbations. The optimization scheme aims to maximize the posterior probability values corresponding to the correct class under Bayesian inference, thereby mitigating significant disruptions in the relative posterior probabilities under attack. The classifier trained using this methodology is denoted as the \textbf{Truth Maximized Diffusion Classifier (TMDC)}.

Our study focuses on investigating the adversarial robustness of the Diffusion Classifier, a generative classifier based on the diffusion model. We propose the Truth Maximization approach to bolster the Diffusion Classifier's robustness against adversarial attacks through training. We conducted comparative analyses between the Diffusion Classifier and TMDC against other commonly utilized neural network classifiers, assessing their resilience under strong adaptive combined attacks and classical white-box attacks. Experimental findings  demonstrate the exceptional adversarial robustness of the Diffusion Classifier relative to alternative classifiers. Furthermore, the efficacy of the Truth Maximization optimization method is confirmed. The optimized classifier, TMDC, achieves remarkable testing accuracies of 82.81\% ($l_{\infty}$) and 86.05\% ($l_{2}$) on the CIFAR-10 dataset under robust Auto Attack \cite{croce2020reliable} settings with parameters set to $\epsilon$=0.05 and version set to ``plus'', thereby attaining the current state-of-the-art performance level. The code for our work is available on GitHub \cite{ourcode}. 

\section{Related Works}

\subsection{Adversarial Attack and Defense}

Since the breakthrough success of AlexNet in 2012 \cite{krizhevsky2012imagenet}, deep neural networks (DNNs) have become pivotal in the realm of computer vision research and application. Subsequent advancements, exemplified by models such as VGG \cite{simonyan2014very}, ResNet \cite{he2016deep}, ViT \cite{dosovitskiy2020image}, and their numerous variants have significantly advanced the state-of-the-art in image classification tasks across prominent datasets. However, despite their outstanding performance in conventional tasks, these models are highly vulnerable to adversarial attacks – techniques devised to mislead deep learning models by introducing imperceptible perturbations to natural data. To assess the robustness of these models, numerous adversarial attack methods have been proposed by previous researchers under both black-box and white-box paradigms \cite{croce2020reliable, croce2020minimally, goodfellow2014explaining, madry2017towards, andriushchenko2020square, shafahi2019adversarial}, with the aim of effectively compromising neural networks. Common strategies to bolster models against such attacks include adversarial training \cite{madry2017towards}, which involves incorporating adversarial perturbations into the training data to improve the model's performance under adversarial conditions. Additionally, methods such as adversarial purification \cite{nie2022diffusion, wang2022guided} have recently gained widespread attention. This approach, focusing on data rather than the model, mitigates adversarial attacks by adding noise into adversarial samples and subsequently denoising them. Nonetheless, such processes may introduce gradient obfuscation issues \cite{athalye2018obfuscated}.

\subsection{Generative Classifiers}
Diverging from discriminative methods that directly delineate data boundaries for image classification, generative approaches, akin to Naive Bayes, first learn the distribution characteristics of image data and then address classification tasks through maximum likelihood estimation modeling. Models such as Naive Bayes \cite{ng2001discriminative}, Energy-Based Models (EBM) \cite{lecun2007energy,lecun2006tutorial}, and the Diffusion Classifier \cite{li2023your, chen2023robust} are constructed under the generative paradigm. Taking Naive Bayes as an example, it models the input image x and label y to derive the data likelihood $p(x|y)$, thereby accomplishing classification through maximum likelihood estimation to derive $p(y|x)$.
\begin{equation}
\label{eq1}
    p\left(y_{i} \mid x\right)=\frac{p\left(y_{i}\right) p\left(x \mid y_{i}\right)}{\sum_{j} p\left(y_{j}\right) p\left(x \mid y_{j}\right)}
\end{equation}
Joint Energy Model (JEM) \cite{grathwohl2019your}, utilizing EBM, reinterprets the standard discriminative classifier of $p(y|x)$ as the joint distribution $p(x,y)$, thereby computing $p(x)$ and $p(x|y)$ to resolve classification tasks. The Diffusion Classifier \cite{li2023your, chen2023robust} simulates data distribution during noise addition and denoising processes, modeling $p(y|x)$ for image classification by maximizing the Evidence Lower Bound (ELBO) of the log-likelihood \cite{blei2017variational}. Previous research has already demonstrated the zero-shot classification capability of the Diffusion Classifier, while our work further showcases its adversarial robustness against adversarial attacks. In this realm, previous attempts have been made, with RDC\cite{chen2023robust} focusing on utilizing the $l$-norm to constrain the input $x$ before and after introducing perturbations to optimize model robustness. In contrast, our work starts from model fine-tuning to enable the model to emerge with the ability to model data under adversarial perturbations.

\subsection{Fine-tuning of Stable Diffusion}
As a powerful and widely acclaimed Text-to-Image image generation model, the Stable Diffusion series \cite{ronneberger2015u,vaswani2017attention,rombach2022high} is often employed directly for tasks such as image classification and image generation. Moreover, fine-tuning the model parameters towards specific image-text pairs for downstream tasks can yield enhanced performance. However, full-parameter fine-tuning of Stable Diffusion poses challenges such as computational resources constraints, time overhead, and potential catastrophic forgetting. In the domain of large language models, the Lora method \cite{hu2021lora,xFormers2022,roich2022pivotal,gal2022image} proposed for Transformer architectures \cite{vaswani2017attention} is suitable for application to Stable Diffusion.

\begin{figure}
    \centering
    \includegraphics[width=0.8\linewidth]{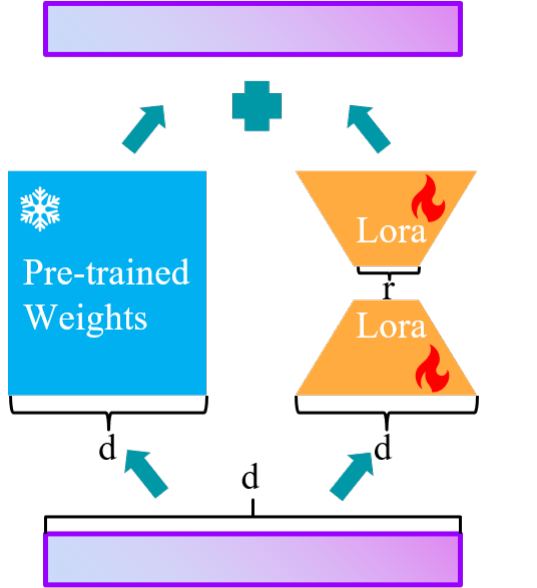}
    \caption{Simplified Illustration of Lora. \textnormal{\textit{Utilizing low-dimensional matrices to approximate high-dimensional ones, where pre-trained weights are frozen, and Lora tensors are employed for training. The memory require during training approaches that of the model's inference process. This configuration reduces both training time and memory overhead, while effectively mitigating catastrophic forgetting.}}}
    \label{fig:Lora}
\end{figure}

\begin{figure*}
    \centering
    \includegraphics[width=1.0\linewidth]{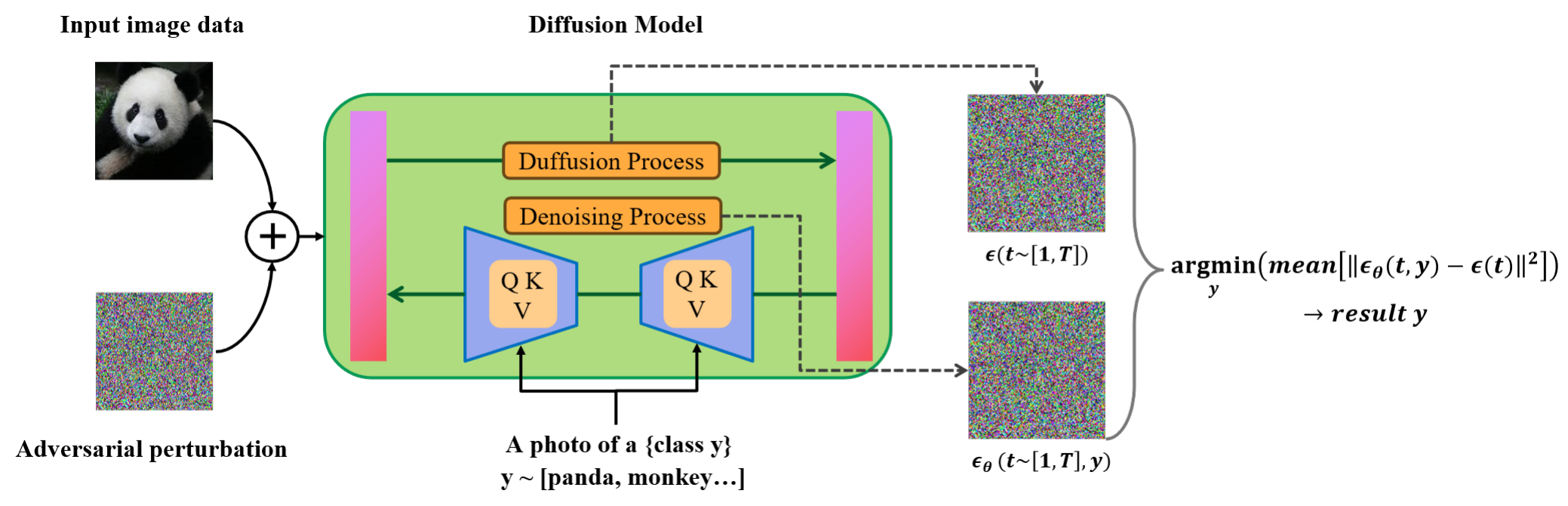}
    \caption{Overview of the Inference Process of the Diffusion Classifier. \textnormal{\textit{Perturbed images are fed into the Diffusion model for both forward noisy processing and backward denoising, with the guiding textual prompt also inputted into the model. The model computes the posterior probabilities corresponding to each class label using Bayes' theorem, and the maximum posterior probability corresponds to the inference result of the classifier. The objective of the inference process in classification can be transformed into selecting the class corresponding to the minimum average error between the noise inferred by the diffusion model at each sampling point and the ground truth value.}}}
    \label{fig:framework}
\end{figure*}

The LoRA method acknowledges that only a small subset of model parameters plays a significant role when targeting specific tasks. Consequently, it becomes feasible to notably diminish the number of training parameters by substituting the high-dimensional parameter matrix with a low-dimensional decomposition matrix. If the size of the pre-trained parameter matrix is set to $d\times d$, it is then replaced with two matrices of size $d\times r$ and $r\times d$ ($d\gg r$), as illustrated in Figure \ref{fig:Lora}. During LoRA fine-tuning, the pre-trained parameters are frozen while the LoRA module undergoes training. Upon completion of training, the Lora parameters are seamlessly integrated with the original parameters, thereby substantially reducing the number of parameters trained during fine-tuning without altering the original parameters. Fine-tuning Stable Diffusion using the LoRA method can drastically reduce training time and significantly alleviate memory requirements.

\section{Methods}
We adopt the method outlined in the Diffusion Classifier \cite{li2023your} to compute class conditional estimates of images utilizing a pre-trained Stable Diffusion model \cite{rombach2022high,https://doi.org/10.48550/arxiv.2204.11824}, thereby constructing an image classifier based on the Diffusion Model for the task of image classification with adversarial perturbations. Subsequently, we propose an approach aimed at enhancing the adversarial robustness of the Diffusion Classifier. \S \ref{sec:3.1} provides an overview of the Diffusion Model, while \S \ref{sec:3.2} outlines the approach of leveraging the Diffusion Model for image classification tasks, with an elaboration on improving its adversarial robustness in \S \ref{sec:3.3}.

\subsection{Diffusion Models}
\label{sec:3.1}
Diffusion models \cite{ho2020denoising} represent a class of discrete-time generative model based on Markov chains. The overall process of the model entails both forward noisy passage and backward denoising. Given an input $x_{0}$, the model performs $T$ rounds of noise addition. Each round of noise addition, denoted as $q\left ( x_{t}\mid x_{t-1} \right ) $, follows a Gaussian distribution, ultimately yielding $x_{T}\sim N(0, I)$. During the denoising process, the model learns the noise added in each round to denoise the image back to $x_{0}$, optionally utilizing low-dimensional text embeddings $y$ for guidance. The denoising process can be represented as $q\left ( x_{t-1}\mid x_{t}, y\right ) $. The entire process can be represented as follows:
\begin{equation}
\label{eq2}
    p_{\theta}\left(\mathbf{x}_{0},y\right)=p\left(\mathbf{x}_{T},y\right) \prod_{t=1}^{T} p_{\theta}\left(\mathbf{x}_{t-1} \mid \mathbf{x}_{t}, y\right)
\end{equation}

Due to the presence of integrals, directly maximizing $p\left(x_{0}\right)$ poses significant challenges. Therefore, the objective is transformed into minimizing the ELBO of the log-likelihood value \cite{blei2017variational}.
\begin{equation}
\label{eq3}
    \log p_{\theta}(x,y)\ge -\mathbb{E}_{\boldsymbol{\epsilon}, t}\left[w_{t}\left\|\boldsymbol{\epsilon}_{\theta}\left(\mathbf{x}_{t}, t,y\right)-\boldsymbol{\epsilon}\right\|_{2}^{2}\right]+C
\end{equation}

We consider $x_{t}=\sqrt{\bar{\alpha}_{t_{i}}} x+\sqrt{1-\bar{\alpha}_{t_{i}}} \epsilon_{i}$, $\mathbb{E}_{\boldsymbol{\epsilon}, t}\left[w_{t}\left\|\boldsymbol{\epsilon}_{\theta}\left(\mathbf{x}_{t}, t\right)-\boldsymbol{\epsilon}\right\|_{2}^{2}\right]$ refers to as diffusion loss in prior studies \cite{kingma2021variational}, and $\epsilon$ follows the standard normal distribution $N\left(0,I\right)$. Previous work has demonstrated that $C$ is typically a negligible value, which can be disregarded in practical computations \cite{li2023your,song2020score}, and in practice, researchers \cite{li2023your,ho2020denoising} often eliminate $W{t}$ to enhance model performance. Thus, we set $W_{t} =1$.


In this transformation, we parameterize the Gaussian noise added at each time step, enabling the neural network to predict the noise at each step during the backward denoising process. The Stable diffusion 2.0 we adopt allows for the selective addition of a text prompt, whose low-dimensional embeddings obtained after text encoding can serve as conditional guidance for the denoising process of the neural network.

\subsection{Diffusion Classifier}
\label{sec:3.2}
In contemporary computer vision literature, prevalent neural network architectures such as Convolutional Neural Networks (CNNs) \cite{lin2017feature,krizhevsky2012imagenet} and Transformer-based architectures \cite{steiner2021train,dosovitskiy2020image} typically adopt discriminative approaches for visual classification tasks. These approaches directly delineate the boundaries of different categories of image data through learning. Conversely, the Diffusion model falls within the realm of generative models. When employed as a classifier, it naturally necessitates the utilization of Bayesian theorem. Specifically, it involves calculating the posterior probability given labels $y$ and modeling of the data $p\left(x\mid y\right)$:
\begin{equation}
\label{eq4}
    p_{\theta}\left(y_{i} \mid x\right)=\frac{p\left(y_{i}\right) p_{\theta}\left(x \mid y_{i}\right)}{\sum_{j} p\left(y_{j}\right) p_{\theta}\left(x \mid y_{j}\right)}
\end{equation}
In the classification process, posterior probabilities corresponding to each class label are computed separately. Therefore, $p\left( y_{i}\right)$ is always equal to $\frac{1}{C} $ (where $C$ represents the total number of classes), allowing for the elimination of $p\left( y_{i}\right)$ during calculations.
\begin{equation}
\label{eq5}
    p_{\theta}\left(y_{i} \mid x\right)=\frac{p_{\theta}\left(x \mid y_{i}\right)}{\sum_{j} p_{\theta}\left(x \mid y_{j}\right)}
\end{equation}
Considering the computational difficulty of $p_{\theta}\left(x\mid y_{i}\right)$, we substitute it with $log\left(p\left(x\mid y_{i}\right)\right)$. Based upon the derivation of the Evidence Lower Bound, 
we combine Eq. \ref{eq5} with Eq. \ref{eq3} to deduce the formula for posterior probability for each class.
\begin{equation}
\label{eq6}
    p_{\theta}\left(y_{i} \mid x\right)=\frac{\exp \left\{-\mathbb{E}_{t, \epsilon}\left[\left\|\epsilon-\epsilon_{\theta}\left(x_{t}, y_{i}\right)\right\|^{2}\right]\right\}}{\sum_{j} \exp \left\{-\mathbb{E}_{t, \epsilon}\left[\left\|\epsilon-\epsilon_{\theta}\left(x_{t}, y_{j}\right)\right\|^{2}\right]\right\}}
\end{equation}
Leveraging the Diffusion model, we can compute the $\epsilon$ for each $t_{i}$ (with the default setting of $i\in \left [ 1,...,1000 \right ]$). Consequently, we can derive unbiased Monte Carlo estimates of the expected value for each class, thus yielding the diffusion loss.
\begin{equation}
\label{eq7}
    mean\left\|\epsilon_{i}-\epsilon_{\theta}\left(\sqrt{\bar{\alpha}_{t_{i}}} x+\sqrt{1-\bar{\alpha}_{t_{i}}} \epsilon_{i}, y\right)\right\|^{2}
\end{equation}
Combining it with the aforementioned derivations, as depicted in Figure \ref{fig:framework}, construction of the generative classification model utilizing the diffusion model is achieved, building upon the work by Li et al. \cite{li2023your}. The work demonstrated the remarkable zero-shot performance of the Diffusion Classifier in open-domain classification scenarios without requiring training. In contrast, our study shifts focus towards its adversarial robustness, utilizing Stable Diffusion 2.0. We contend that it exhibits superior resilience against adversarial perturbations in images without requiring training, compared to other neural networks.
\subsection{Robust Truth Maximization}
\label{sec:3.3}
After conducting comparative experiments under various attacks, we have demonstrated the adversarial robustness of the Diffusion Classifier. Furthermore, we delve into strategies to enhance its robustness, aiming to contribute more to the research on robustness of classification models. To enhance the classifier's accuracy, according to Eq. \ref{eq6}, the model should be trained to minimize its diffusion loss, $\mathbb{E}_{\boldsymbol{\epsilon}, t}\left[w_{t}\left\|\boldsymbol{\epsilon}_{\theta}\left(\mathbf{x}_{t}, t\right)-\boldsymbol{\epsilon}\right\|_{2}^{2}\right]$, when provided with the ground-truth class labels as input. This entails shifting the model's backward denoising inference values, guided by the true labels, towards the ground truth values.

In order to enhance the robustness of diffusion model against adversarial attacks, we draw inspiration from the traditional adversarial training employed in vision classifiers \cite{madry2017towards}. While generative models cannot directly model the data boundaries between different classes during adversarial sample training, optimizing the model by inputting adversarial samples along with their ground-truth labels and minimizing diffusion loss can improve the model's capability to model samples augmented by adversarial attacks. Following Eq.\ref{eq6} and Eq.\ref{eq7}, we define the training loss as:
\begin{equation}
\label{eq8}
    Loss=\frac{1}{T}\sum_{t=0}^{T-1}\left[\left\|\epsilon_{\theta}\left ( t,y_{true}\right )-\epsilon \left (t \right ) \right\|^{2}\right]
\end{equation}

Our work utilizes pre-trained Stable Diffusion 2.0 with approximately 354 million parameters. Performing full-parameter training would incur significant memory and time overheads, potentially compromising the pre-trained model's image modeling capabilities. Thus, we employ the Lora fine-tuning technique to mitigate this issue. By employing a decomposition method that approximates high-dimensional parameter matrices with low-dimensional matrices, we reduce the memory requirements for training to the level of model inference. The trained Lora module is then seamlessly merged with the pre-trained parameters to maintain the original modeling capabilities of the pre-trained model.

During the training process, we input augmented samples $x$ from the training set along with their correct labels $y$ into Stable Diffusion. Then a pre-trained scheduler is employed for noise injection, and the model predict the noise at each time step, calculate the loss, and then minimize it. The model obtained through this approach is referred to as the Truth Maximized Diffusion Classifier (TMDC) by us. For a detailed outline of the classifier's training and inference process, please refer to Algorithm \ref{alg: alg1}.

\begin{algorithm}[h]
    \caption{Truth Maximized Diffusion Classifier(\textbf{TMDC})}
    \textbf{Notation:} \(X\): dataset, \(N\): data batch, $x$: image, $y$: ground-truth label, \(\epsilon\): model prediction, \(\tau\): learning rate, \(T\): time step, $W$: weights of diffusion model, $L$: List of data class(car, truck, horse, ...,  plane)\\
    \textbf{Model Training}\\
    \label{alg: alg1}
    \begin{algorithmic}[1]
        \FOR{\( N\in X\)}
        \STATE $x, y \longleftarrow N$
        \FOR{$t$ in $T$}
        \STATE $\epsilon\left(t \right) \longleftarrow scheduler\left(x,t \right)$
        \ENDFOR
        \FOR{$t$ in $T$}
        \STATE $\epsilon_{\theta}\left( t, y\right) \longleftarrow model\ predict\left( x,y,t\right)$
        \ENDFOR
        \STATE $Loss \longleftarrow \frac{1}{T}\sum_{t=0}^{T-1}\left[\left\|\epsilon_{\theta}\left ( t,y_{true}\right )-\epsilon \left (t \right ) \right\|^{2}\right]$
        \STATE $g \longleftarrow \nabla Loss$
        \STATE $W \longleftarrow W-\tau g$
        \ENDFOR
        \RETURN W
    \end{algorithmic}
    \textbf{Model Inference}\\
    \begin{algorithmic}[1]
        \FOR{$N \in X$}
            \FOR{$y \in L$}
                \STATE $LossList\left[ y \right] \longleftarrow list\left( \right)$
                \FOR{$t$ in $T$}
                    \STATE $\epsilon\left(t \right) \longleftarrow scheduler\left(x,t \right)$
                    \STATE $\epsilon_{\theta}\left( t, y\right) \longleftarrow model\ predict\left( x,y,t\right)$
                    \STATE $LossList.append\left( \left\|\epsilon_{\theta}\left ( t,y\right )-\epsilon \left (t \right ) \right\|^{2} \right)$
                \ENDFOR
                \STATE $LossList\left[y \right] \longleftarrow mean\left( LossList\left[y \right] \right)$
            \ENDFOR
            \STATE $result \longleftarrow \underset{y \in L}{\arg \min } \left(\operatorname{Errors}\left[y\right]\right)$
        \ENDFOR
        \RETURN result
    \end{algorithmic}
\end{algorithm}

\section{Experiments}
We conducted a series of rigorous experiments, employing various black-box and white-box attack methods to assess the adversarial robustness of both the Diffusion Classifier and TMDC. Furthermore, we compared their performance with popular neural networks in the field of computer vision. \S \ref{sec:4.1} elucidates the detailed experimental setup and training specifics of the models. \S \ref{sec:4.2} showcases the results of the robustness study of the Diffusion Classifier under several classical white-box attacks. \S \ref{sec:4.3} presents the model's performance under Auto Attack, a widely recognized black-box and white-box combined attack method. Lastly, \S \ref{sec:4.4} entails the ablation study of the TMDC method.

\subsection{Experiment Settings}
\label{sec:4.1}
\textbf{Dataset:} Considering the characteristics of the dataset and the time overhead incurred by attack algorithms and model training, we opt for CIFAR10 \cite{krizhevsky2009learning} to conduct our experiments. To assess the adversarial robustness of Diffusion Classifier and TMDC, inspired by the method of utilizing a subset of data for detection as proposed in DiffPure \cite{nie2022diffusion}, and to eliminate testing randomness, we select 1024 data points from the CIFAR10 test set of 10,000 items for evaluation. Moreover, during the training process of the TMDC method, we endeavor to optimize Stable Diffusion 2.0 on the CIFAR10 training set.

\noindent\textbf{Practical Implementation Setup:} In the naive implementation process of Algorithm \ref{alg: alg1} for model inference, it necessitates computing over all time steps for each class in the category list for classification. This inevitably imposes a heavy computational burden. Inspired by the upper confidence bound algorithm \cite{krizhevsky2009learning}, it is possible to save computation by prematurely discarding class labels that significantly fail to meet classification requirements based on diffusion loss. When dealing with CIFAR10, we adhere to the setup proposed by Li et al. \cite{li2023your}, where we initially compute losses for all labels over 50 time steps, discard the top 5 labels with the highest losses, and proceed with computations over 500 time steps for the remaining labels, thereby obtaining the final classification results.

\noindent\textbf{Training Setup:} We conducted training of the diffusion model on a single A100 (80GB) GPU, with a batch size set to 4. We employed the AdamW optimizer with a learning rate of 1e-6, beta parameters set to (0.9, 0.999), weight decay of 1e-2, and epsilon set to 1e-8. Optimization was performed over 3,000 steps on the CIFAR10 training set, utilizing a constant with warmup learning rate scheduler with a warmup step of 100. For the final experimental evaluation, we selected the checkpoint after 200 steps of optimization, a configuration validated in the ablation study of \S \ref{sec:4.4}.

\subsection{White-box Attack Robustness}
\label{sec:4.2}
\begin{table*}[t]
\caption{Comparison of White-box Adversarial Robustness. \textnormal{\textit{ResNet50, Vit\_B/16, and WideResNet50 were all trained on the CIFAR10 dataset, and then subjected to robustness testing on test data with adversarial perturbations. In contrast, the Diffusion Classifier was directly tested on the test set with added attacks.}}}
\label{tab:1}
\centering
\renewcommand{\arraystretch}{1.5} \resizebox{0.7\textwidth}{!}
{
\begin{tabular}{lccc}
\hline
{\color[HTML]{333333} \textbf{baselines}}                           & {\color[HTML]{333333} \textbf{Clean}}            & {\color[HTML]{333333} \textbf{FGSM$\left(\epsilon=0.05\right)$}}             & {\color[HTML]{333333} \textbf{PGD$\left(\epsilon=0.05,iter=40 \right)$}}              \\ \hline
{\color[HTML]{333333} ResNet50 \cite{he2016deep}}                            & {\color[HTML]{333333} 90.51\%}          & {\color[HTML]{333333} 39.77\%}          & {\color[HTML]{333333} 0.0\%}            \\
{\color[HTML]{333333} Vit\_B/16 \cite{dosovitskiy2020image}}                           & {\color[HTML]{333333} 98.10\%}          & {\color[HTML]{333333} 23.69\%}          & {\color[HTML]{333333} 0.0\%}            \\
{\color[HTML]{333333} WideResNet50 \cite{zagoruyko2016wide}}                        & {\color[HTML]{333333} 98.05\%}          & {\color[HTML]{333333} 22.40\%}           & {\color[HTML]{333333} 0.0\%}            \\
{\color[HTML]{333333} \textbf{Diffusion Classifier(OURS)}} & {\color[HTML]{333333} 89.44\%} & {\color[HTML]{333333} \textbf{50.17\%}} & {\color[HTML]{333333} \textbf{42.30\%}} \\ \hline
\end{tabular}}
\end{table*}

In this section, we employ two widely-used white-box attack algorithms to introduce adversarial perturbations to the test data, thereby evaluating the adversarial robustness of the Diffusion Classifier, shown in \S \ref{sec:4.2.1}. Additionally, we subject TMDC to attacks of the same intensity. In contrast, the remaining models undergo adversarial training for comparison, aiming to assess the effectiveness of our model optimization compared to the widely-used adversarial training on discriminative classifiers, as described in \S \ref{sec:4.2.2}.

\noindent\textbf{Adversarial Attacks:} This experiment employs two white-box attack algorithms, namely FGSM \cite{goodfellow2014explaining} and PGD \cite{madry2017towards}. We use a pre-trained ResNet50 model on CIFAR10 as the attack generator to introduce perturbations into the test data. The FGSM algorithm, contrary to the gradient descent method used in neural network training optimization, adds smooth perturbations, denoted as $\epsilon$, along the direction of the gradient constrained by the $l_{\infty}-norm$ to maximize the loss function, leading to misclassification by the model. The PGD attack method is an improved version of FGSM, performing multiple iterations based on single-step attacks under the $l_{\infty}-norm$ to achieve better attack effectiveness. In our experiment, we set the $\epsilon$ parameter for FGSM and PGD attacks to 0.05, while the number of iterations for PGD attacks is set to 40.

\subsubsection{\textbf{Comparison of White-box Adversarial Robustness}}
\label{sec:4.2.1}
We extracted 1024 samples from the CIFAR10 dataset and introduced adversarial perturbations using FGSM and PGD. Then, we utilized a rapid algorithm for staged label elimination to allow the classifier to infer predicted labels and calculate the model's accuracy under adversarial attacks, in conjunction with the ground-truth labels, for comparison with other popular neural networks trained on CIFAR10. The experimental results are presented in Table \ref{tab:1}.

Under the FGSM attack, the accuracy of ResNet50 dropped from 90.51\% to 39.77\%, Vit\_B/16 from 98.10\% to 23.69\%, and WideResNet50 from 98.05\% to 22.40\%, all experiencing decreases of over 50\%. Vit and WideResNet50 even dropped by over 75\%. Conversely, the untrained Diffusion Classifier achieved an accuracy of 89.44\% on the clean data, dropping to 50.17\% under the FGSM attack, with a significantly lower decrease in accuracy compared to other models. Under the PGD attack algorithm with $\epsilon$ set to 0.05 and 40 iterations, the accuracy of all other models dropped to 0.0\%, whereas that of the Diffusion Classifier only decreased to 42.30\%. These experimental results demonstrate the outstanding robustness of the Diffusion Classifier against white-box adversarial attacks when compared to other neural networks, even in an untrained state.

\subsubsection{\textbf{Comparison between Truth Maximization and Adversarial Training}}
\label{sec:4.2.2}

\begin{table}[t]
\caption{Comparison between Truth Maximization and Adversarial Training. \textnormal{\textit{ResNet50, Vit\_B/16, and WideResNet50 all underwent multiple rounds of adversarial training on data augmented with the PGD algorithm. Training was halted once the model's classification performance stabilized, after which robustness testing was conducted using the test set. In contrast, the Diffusion Classifier was subjected to robustness testing after optimization through Truth Maximization on the same augmented data.}}} 
\label{tab:2}
\centering
\renewcommand{\arraystretch}{1.5} \resizebox{0.35\textwidth}{!}{
\begin{tabular}{lc}
\hline
\textbf{Baselines}    & \textbf{PGD$\left(\epsilon=0.05,iter=40 \right)$}     \\ \hline
ResNet50     & 45.39\% \\
Vit\_B/16    & 39.72\% \\
WideResNet50 & 45.77\% \\
\textbf{TMDC(OURS)}   & \textbf{70.02\%} \\ \hline
\end{tabular}}
\end{table}

We trained Stable Diffusion using data augmented with PGD adversarial perturbations, resulting in the model TMDC. Meanwhile, the other models underwent adversarial training using the PGD algorithm \cite{madry2017towards}. Then we conducted a comparative study of the robustness of each model on the test set under PGD attacks. The experimental results are presented in Table \ref{tab:2}.

After undergoing adversarial training, the accuracy of ResNet50 under PGD attacks increased from its original 0.0\% to 45.39\%, while Vit\_B/16 rose to 39.72\%, and WideResNet50 increased to 45.77\%. This demonstrates that adversarial training can effectively enhance the adversarial robustness of widely used discriminative classifiers. Meanwhile, TMDC achieved an accuracy of 70.02\% under the same adversarial attacks, significantly outperforming the commonly used adversarial training methods in enhancing model robustness. Thus, compared to discriminative classifiers, which can conveniently improve robustness through adversarial training, Diffusion Classifier as a generative classifier also possesses meaningful Truth Maximization optimization methods.

\subsection{Auto Attack Robustness}
\label{sec:4.3}

\begin{table*}[]
\caption{Comparison of Auto Attack Robustness. \textnormal{\textit{Under both $L_{\infty}$ and $L_{2}$ norm constraints, Auto Attack was conducted with epsilon set to 0.05 and seed set to 2024. In the experiments involving DiffPure, the samples processed through DiffPure were re-fed into ResNet50, which was pre-trained on CIFAR10, for testing. And the ResNet50 model used in this experiment shares the same weight values as the models in the comparative experiment.}}}
\label{tab:3}
\centering
\renewcommand{\arraystretch}{1.5} \resizebox{0.7\textwidth}{!}{
\begin{tabular}{lcc}
\hline
\textbf{baselines}                  & \textbf{Auto Attack($l_{\infty}-norm$)} & \textbf{Auto Attack($l_{2}-norm$)} \\ \hline
\textbf{DiffPure \cite{nie2022diffusion} }                 & \textbf{57.94\%}          & \textbf{75.34\%}        \\
JEM   \cite{grathwohl2019your}                     & 10.13\%          & 26.56\%        \\
WideResNet50  \cite{zagoruyko2016wide}             & 0.0\%            & 23.98\%        \\
Vit\_B/16     \cite{dosovitskiy2020image}             & 0.0\%            & 31.42\%        \\
ResNet50  \cite{he2016deep}                 & 0.50\%           & 37.52\%        \\
\textbf{Diffusion Classifier(OURS)} & \textbf{79.52}\%          & \textbf{81.18}\%        \\
\textbf{TMDC(OURS)}                 & \textbf{82.81\%}          & \textbf{86.05\%}        \\ \hline
\end{tabular}}
\end{table*}

In this section, we employ the Auto Attack method to evaluate the adversarial robustness of both the Diffusion Classifier and TMDC under combined attacks. For rigorous conclusions, we compare them with discriminative classifiers and introduce the JEM generative classifier for comparative experiments. Furthermore, we incorporate another widely recognized approach for combating adversarial attacks, DiffPure, into our experiments.

\noindent\textbf{Adversarial Attack:} Auto Attack \cite{croce2020reliable} is a combined adversarial attack method that encompasses both black-box and white-box attacks. It improves upon PGD by employing the APGD algorithm, which automatically adjusts the step size. It rapidly moves with a larger step size and gradually reduces the step size to maximize the objective function locally. If the step size halving is detected, it restarts at the local maximum, thereby mounting more effective attacks against neural networks. APGD comes in different versions depending on the target loss function, and Auto Attack combines various versions of APGD along with Square attack (black-box) and FAB attack (white-box), forming a combination of black-box and white-box attacks. In this experiment, we set the version of Auto Attack to ``plus'' (a combination of all types of algorithms) and use both $l_{2}$ and $l_{\infty}$ norms to constrain the perturbation.

Stable Diffusion 2.0 is optimized using the Truth Maximization method on data augmented with Auto Attack. Subsequently, experiments are conducted on the test set augmented with the same attacks to assess its adversarial robustness against Auto Attack. The remaining comparative approaches are subjected to attacks using the same algorithm, with all groups utilizing adversarial samples generated by a pre-trained ResNet50 model on CIFAR10. The experimental results are presented in Table \ref{tab:3}.

Under the $l_{\infty}$ norm-constrained Auto Attack, the accuracy of WideResNet50 and Vit\_B/16 on the test set plummeted to 0.0\%, while that of ResNet50 dropped to 0.5\%. However, after purifying the perturbations through DiffPure, the accuracy of ResNet50 reached 57.94\%. Moreover, JEM achieved an accuracy of 10.13\% under Auto Attack. In contrast, the Diffusion Classifier exhibited excellent robustness against this combination of black-box and white-box attacks, achieving an accuracy of 79.52\% without any training. After optimization using the Truth Maximization method, TMDC's accuracy further increased to 82.81\%.

Meanwhile, under the $l_{2}$ norm-constrained Auto Attack, the accuracy of WideResNet50 on the test set was 23.98\%, while Vit\_B/16 achieved 31.42\%, and ResNet50 reached 37.52\%. However, after purifying the perturbations through DiffPure, the accuracy of ResNet50 rose to 75.34\%. JEM attained an accuracy of 26.56\% under this norm of Auto Attack. The untrained Diffusion Classifier achieved an accuracy of 81.18\%, while TMDC reached 86.05\%. Under both $l_{\infty}$ and $l_{2}$ norm-constrained Auto Attack scenarios, classifiers constructed from Stable Diffusion 2.0 demonstrated superior adversarial robustness compared to other comparative models. Furthermore, compared to the strategy of purifying the data and refeeding it into ResNet50 after DiffPure, the Diffusion Classifier also exhibited higher classification performance.

\subsection{Ablation Study}
\label{sec:4.4}

In order to validate the effectiveness of the Truth Maximization optimization approach employed in our work for the Diffusion Classifier, as well as the correctness of the selection settings for checkpoints during the training process, we conducted an ablation study. \S \ref{sec:4.4.1} presents our investigation into Truth Maximization, while \S \ref{sec:4.4.2} outlines our experiments concerning different checkpoint selection.

\subsubsection{\textbf{Ablation on Truth Maximization}}
\label{sec:4.4.1}

\begin{figure}[t]
    \centering
    \includegraphics[width=1.0\linewidth]{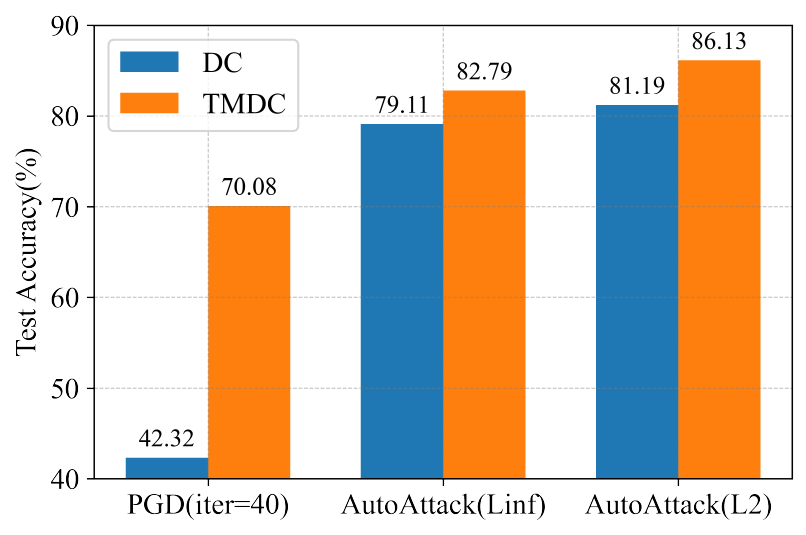}
    \caption{Comparison between Diffusion Classifier and TMDC. \textnormal{\textit{The PGD attack is conducted with parameters set as follows: $\epsilon$ is set to 0.05, and the attack runs for 40 iterations, in accordance with Section 4.2. As for Auto Attack, its version is uniformly designated as ``plus'', with $\epsilon$ set to 0.05 and the seed initialized with five sets of distinct random numbers.}}}
    \label{fig:Ablation1}
\end{figure}

In this experiment, we employ PGD($iter=40$), Auto Attack ($l_{\infty}$), and Auto Attack ($l_{2}$) as three adversarial attack methods. For each attack, we randomly select 5 sets of different seeds to sample data from the test set. We evaluate the accuracy metrics of the Diffusion Classifier and TMDC in performing classification tasks under these attacks, and report the averaged test results. The experimental outcomes are illustrated in Figure \ref{fig:Ablation1}.

Under three types of adversarial attacks, the Truth Maximization approach consistently yields effective optimizations for the Diffusion Classifier. Specifically, under the PGD attack, models optimized through Truth Maximization demonstrate a substantial enhancement in average testing accuracy, soaring from 42.32\% to 70.08\%, marking an increase of 65.59\%. Notably, the model's adversarial robustness under this attack type experiences significant improvement. Moreover, under the two norm-constrained Auto Attack scenarios, the accuracy elevates from 79.11\% ($l_{\infty}$) and 81.19\% ($l_{2}$) to 82.79\% and 86.13\%, respectively, showcasing notable optimizations under combined attacks. Further corroborated by the experimental findings in \S \ref{sec:4.2.2}, TMDC consistently achieves a higher accuracy of 70.02\% under PGD attack compared to other classification models using adversarial training. These experimental outcomes collectively underscore the efficacy of the Truth Maximization methodology in enhancing the adversarial robustness of the Diffusion Classifier. Furthermore, in contrast to adversarial training, applying Truth Maximization to diffusion models exhibits superior performance.

\subsubsection{\textbf{Ablation on Checkpoint Selection}}
\label{sec:4.4.2}

\begin{figure}[t]
    \centering
    \includegraphics[width=\linewidth]{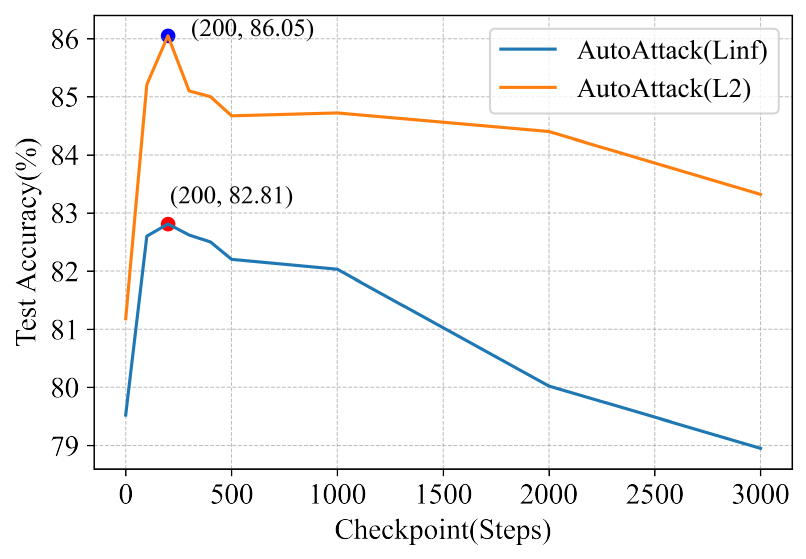}
    \caption{Study on Checkpoint Selection. \textnormal{\textit{For Auto Attack, the version is uniformly set to ``plus'', with a value of 0.05 for parameter $\epsilon$, and the seed is fixed at 2024. Throughout the Truth Maximization training process, a learning rate scheduler employing ``constant with warmup'' strategy is employed, wherein the learning rate is set to 1e-6, and the warm-up steps are configured to be 100. Both sets of experiments undergo optimization for 3000 steps.}}}
    \label{fig:Ablation2}
\end{figure}

In this experiment, we conducted trials under two norm-constrained Auto Attack scenarios. The selection of test dataset for both sets of experiments employed the same random seed. Moreover, all experiments underwent optimization using the Truth Maximization methodology for 3000 steps. During this optimization process, checkpoints were saved every 100 steps for the first 500 steps, followed by checkpoints saved every 1000 steps thereafter. The settings for optimizer and learning rate scheduler remained consistent with those outlined in \S \ref{sec:4.1}, ensuring the validity and coherence of the experimental setup. The experimental results are shown in Figure \ref{fig:Ablation2}.

Truth Maximization is a method employed to enhance the classification capability of the Diffusion Classifier by minimizing diffusion loss, which serves as the objective function, when training the model with both the ground-truth labels of the training set and perturbed images. This approach aims to strengthen the diffusion model's ability to model enhanced images conditioned on the correct labels. However, it does not directly improve the model's capacity to model boundaries between different data types. Therefore, we must consider the impact of optimization steps on the classifier's ultimate performance. As depicted in Figure \ref{fig:Ablation2}, when checkpoints are saved every 100 steps, the model's accuracy on the test set reaches its peak around the 200th step checkpoint, with the test accuracy reaching 86.05\% and 82.81\% respectively, and gradually decreases thereafter. Furthermore, in the experimental group under $l_{\infty}$ norm constraints, the model's accuracy at the 3000th step is even lower than before optimization. After 200 steps of training, the model becomes overfitted to the training data, resulting in weakened classification performance due to blurred diffusion loss boundaries guided by different class labels in the test data. Consequently, after Truth Maximization training, we select the model from the 200th step checkpoint for subsequent testing to achieve relatively fair classifier performance.

\section{Discussion}
\textit{\textbf{Collaboration with Purification:}} The generation of adversarial samples for image classification or adversarial training using diffusion model is subject to uncertainty stemming from shifts in image data distribution, rendering it vulnerable to high-intensity adaptive attacks. This vulnerability is partially attributed to constraints imposed by the performance of classifiers used after generating purified images. However, comparative experiments demonstrate that purification-based methods consistently outperform other baseline approaches. Therefore, anchored in the purification paradigm, developing a classifier based on diffusion models, that leverages the statistical uncertainty of data and utilizes different class posterior probabilities for classification, holds promise for bolstering adversarial resilience.

\noindent\textbf{\textit{Decoupling from Training:}} Despite achieving excellent adversarial robustness, our proposed TMDC method remains constrained by the necessity of training on adversarial samples, requiring a dedicated training set for the diffusion model, thereby posing inefficiencies in terms of computational resources and time. To mitigate these challenges, decoupling from training, segmenting the inference process of the diffusion model into multiple stages, and optimizing the sampling strategy offer a fertile ground for exploration. Such an approach not only enhances the model's classification performance under adversarial attacks but also improves inference efficiency, consequently conserving computational resources and time.

\section{Conclusion}
In light of the widespread vulnerability observed in commonly used visual neural network classifiers when subjected to adversarial attacks, we conducted thorough assessments and found that the Diffusion Classifier, derived from a robust generative model, demonstrates excellent adversarial robustness. Utilizing the diffusion model as a conditional density estimator, we modeled image data guided by text prompts through the combination of Evidence Lower Bound (ELBO) and unbiased Monte Carlo estimation, leveraging Bayesian theorem to construct the classifier. Additionally, we propose a model optimization approach termed Truth Maximization, which, through training guided by ground-truth labels, further enhances the adversarial robustness of the pre-trained Stable Diffusion-based generative classifier. Models trained using this approach are denoted as Truth Maximization Diffusion Classifier(TMDC). Through empirical evaluation against classical white-box attacks and widely employed strong combined adaptive attacks like Auto Attack, we demonstrated the exceptional adversarial robustness of the Diffusion Classifier even in the absence of explicit training. Moreover, the optimized TMDC model achieved state-of-the-art performance against strong white-box attacks and combined adaptive attacks on the CIFAR-10 dataset.


\bibliographystyle{unsrt}
\bibliography{sample-base}

\begin{thebibliography}{10}

\bibitem{deng2009imagenet}
Jia Deng, Wei Dong, Richard Socher, Li-Jia Li, Kai Li, and Li~Fei-Fei.
\newblock Imagenet: A large-scale hierarchical image database.
\newblock In {\em 2009 IEEE conference on computer vision and pattern recognition}, pages 248--255. Ieee, 2009.

\bibitem{krizhevsky2012imagenet}
Alex Krizhevsky, Ilya Sutskever, and Geoffrey~E Hinton.
\newblock Imagenet classification with deep convolutional neural networks.
\newblock {\em Advances in neural information processing systems}, 25, 2012.

\bibitem{lecun1989handwritten}
Yann LeCun, Bernhard Boser, John Denker, Donnie Henderson, Richard Howard, Wayne Hubbard, and Lawrence Jackel.
\newblock Handwritten digit recognition with a back-propagation network.
\newblock {\em Advances in neural information processing systems}, 2, 1989.

\bibitem{szegedy2015going}
Christian Szegedy, Wei Liu, Yangqing Jia, Pierre Sermanet, Scott Reed, Dragomir Anguelov, Dumitru Erhan, Vincent Vanhoucke, and Andrew Rabinovich.
\newblock Going deeper with convolutions.
\newblock In {\em Proceedings of the IEEE conference on computer vision and pattern recognition}, pages 1--9, 2015.

\bibitem{zagoruyko2016wide}
Sergey Zagoruyko and Nikos Komodakis.
\newblock Wide residual networks.
\newblock {\em arXiv preprint arXiv:1605.07146}, 2016.

\bibitem{he2016deep}
Kaiming He, Xiangyu Zhang, Shaoqing Ren, and Jian Sun.
\newblock Deep residual learning for image recognition.
\newblock In {\em Proceedings of the IEEE conference on computer vision and pattern recognition}, pages 770--778, 2016.

\bibitem{vaswani2017attention}
Ashish Vaswani, Noam Shazeer, Niki Parmar, Jakob Uszkoreit, Llion Jones, Aidan~N Gomez, {\L}ukasz Kaiser, and Illia Polosukhin.
\newblock Attention is all you need.
\newblock {\em Advances in neural information processing systems}, 30, 2017.

\bibitem{liu2021swin}
Ze~Liu, Yutong Lin, Yue Cao, Han Hu, Yixuan Wei, Zheng Zhang, Stephen Lin, and Baining Guo.
\newblock Swin transformer: Hierarchical vision transformer using shifted windows.
\newblock In {\em Proceedings of the IEEE/CVF international conference on computer vision}, pages 10012--10022, 2021.

\bibitem{dosovitskiy2020image}
Alexey Dosovitskiy, Lucas Beyer, Alexander Kolesnikov, Dirk Weissenborn, Xiaohua Zhai, Thomas Unterthiner, Mostafa Dehghani, Matthias Minderer, Georg Heigold, Sylvain Gelly, et~al.
\newblock An image is worth 16x16 words: Transformers for image recognition at scale.
\newblock {\em arXiv preprint arXiv:2010.11929}, 2020.

\bibitem{kim2022adaface}
Minchul Kim, Anil~K Jain, and Xiaoming Liu.
\newblock Adaface: Quality adaptive margin for face recognition.
\newblock In {\em Proceedings of the IEEE/CVF conference on computer vision and pattern recognition}, pages 18750--18759, 2022.

\bibitem{an2022killing}
Xiang An, Jiankang Deng, Jia Guo, Ziyong Feng, XuHan Zhu, Jing Yang, and Tongliang Liu.
\newblock Killing two birds with one stone: Efficient and robust training of face recognition cnns by partial fc.
\newblock In {\em Proceedings of the IEEE/CVF Conference on Computer Vision and Pattern Recognition}, pages 4042--4051, 2022.

\bibitem{hashimoto2020multi}
Noriaki Hashimoto, Daisuke Fukushima, Ryoichi Koga, Yusuke Takagi, Kaho Ko, Kei Kohno, Masato Nakaguro, Shigeo Nakamura, Hidekata Hontani, and Ichiro Takeuchi.
\newblock Multi-scale domain-adversarial multiple-instance cnn for cancer subtype classification with unannotated histopathological images.
\newblock In {\em Proceedings of the IEEE/CVF conference on computer vision and pattern recognition}, pages 3852--3861, 2020.

\bibitem{maksoud2020sos}
Sam Maksoud, Kun Zhao, Peter Hobson, Anthony Jennings, and Brian~C Lovell.
\newblock Sos: Selective objective switch for rapid immunofluorescence whole slide image classification.
\newblock In {\em Proceedings of the IEEE/CVF Conference on Computer Vision and Pattern Recognition}, pages 3862--3871, 2020.

\bibitem{xu2023deep}
Wenjia Xu, Jiuniu Wang, Zhiwei Wei, Mugen Peng, and Yirong Wu.
\newblock Deep semantic-visual alignment for zero-shot remote sensing image scene classification.
\newblock {\em ISPRS Journal of Photogrammetry and Remote Sensing}, 198:140--152, 2023.

\bibitem{kuckreja2023geochat}
Kartik Kuckreja, Muhammad~Sohail Danish, Muzammal Naseer, Abhijit Das, Salman Khan, and Fahad~Shahbaz Khan.
\newblock Geochat: Grounded large vision-language model for remote sensing.
\newblock {\em arXiv preprint arXiv:2311.15826}, 2023.

\bibitem{blei2017variational}
David~M Blei, Alp Kucukelbir, and Jon~D McAuliffe.
\newblock Variational inference: A review for statisticians.
\newblock {\em Journal of the American statistical Association}, 112(518):859--877, 2017.

\bibitem{li2023your}
Alexander~C Li, Mihir Prabhudesai, Shivam Duggal, Ellis Brown, and Deepak Pathak.
\newblock Your diffusion model is secretly a zero-shot classifier.
\newblock In {\em Proceedings of the IEEE/CVF International Conference on Computer Vision}, pages 2206--2217, 2023.

\bibitem{croce2020reliable}
Francesco Croce and Matthias Hein.
\newblock Reliable evaluation of adversarial robustness with an ensemble of diverse parameter-free attacks.
\newblock In {\em International conference on machine learning}, pages 2206--2216. PMLR, 2020.

\bibitem{ourcode}
Tmdc.
\newblock \url{https://github.com/zzhjsaiosjd/TMDC_SD_2}.

\bibitem{simonyan2014very}
Karen Simonyan and Andrew Zisserman.
\newblock Very deep convolutional networks for large-scale image recognition.
\newblock {\em arXiv preprint arXiv:1409.1556}, 2014.

\bibitem{croce2020minimally}
Francesco Croce and Matthias Hein.
\newblock Minimally distorted adversarial examples with a fast adaptive boundary attack.
\newblock In {\em International Conference on Machine Learning}, pages 2196--2205. PMLR, 2020.

\bibitem{goodfellow2014explaining}
Ian~J Goodfellow, Jonathon Shlens, and Christian Szegedy.
\newblock Explaining and harnessing adversarial examples.
\newblock {\em arXiv preprint arXiv:1412.6572}, 2014.

\bibitem{madry2017towards}
Aleksander Madry, Aleksandar Makelov, Ludwig Schmidt, Dimitris Tsipras, and Adrian Vladu.
\newblock Towards deep learning models resistant to adversarial attacks.
\newblock {\em arXiv preprint arXiv:1706.06083}, 2017.

\bibitem{andriushchenko2020square}
Maksym Andriushchenko, Francesco Croce, Nicolas Flammarion, and Matthias Hein.
\newblock Square attack: a query-efficient black-box adversarial attack via random search.
\newblock In {\em European conference on computer vision}, pages 484--501. Springer, 2020.

\bibitem{shafahi2019adversarial}
Ali Shafahi, Mahyar Najibi, Mohammad~Amin Ghiasi, Zheng Xu, John Dickerson, Christoph Studer, Larry~S Davis, Gavin Taylor, and Tom Goldstein.
\newblock Adversarial training for free!
\newblock {\em Advances in neural information processing systems}, 32, 2019.

\bibitem{nie2022diffusion}
Weili Nie, Brandon Guo, Yujia Huang, Chaowei Xiao, Arash Vahdat, and Anima Anandkumar.
\newblock Diffusion models for adversarial purification.
\newblock {\em arXiv preprint arXiv:2205.07460}, 2022.

\bibitem{wang2022guided}
Jinyi Wang, Zhaoyang Lyu, Dahua Lin, Bo~Dai, and Hongfei Fu.
\newblock Guided diffusion model for adversarial purification.
\newblock {\em arXiv preprint arXiv:2205.14969}, 2022.

\bibitem{athalye2018obfuscated}
Anish Athalye, Nicholas Carlini, and David Wagner.
\newblock Obfuscated gradients give a false sense of security: Circumventing defenses to adversarial examples.
\newblock In {\em International conference on machine learning}, pages 274--283. PMLR, 2018.

\bibitem{ng2001discriminative}
Andrew Ng and Michael Jordan.
\newblock On discriminative vs. generative classifiers: A comparison of logistic regression and naive bayes.
\newblock {\em Advances in neural information processing systems}, 14, 2001.

\bibitem{lecun2007energy}
Yann LeCun, Sumit Chopra, M~Ranzato, and F-J Huang.
\newblock Energy-based models in document recognition and computer vision.
\newblock In {\em Ninth International Conference on Document Analysis and Recognition (ICDAR 2007)}, volume~1, pages 337--341. IEEE, 2007.

\bibitem{lecun2006tutorial}
Yann LeCun, Sumit Chopra, Raia Hadsell, M~Ranzato, and Fujie Huang.
\newblock A tutorial on energy-based learning.
\newblock {\em Predicting structured data}, 1(0), 2006.

\bibitem{chen2023robust}
Huanran Chen, Yinpeng Dong, Zhengyi Wang, Xiao Yang, Chengqi Duan, Hang Su, and Jun Zhu.
\newblock Robust classification via a single diffusion model.
\newblock {\em arXiv preprint arXiv:2305.15241}, 2023.

\bibitem{grathwohl2019your}
Will Grathwohl, Kuan-Chieh Wang, J{\"o}rn-Henrik Jacobsen, David Duvenaud, Mohammad Norouzi, and Kevin Swersky.
\newblock Your classifier is secretly an energy based model and you should treat it like one.
\newblock {\em arXiv preprint arXiv:1912.03263}, 2019.

\bibitem{ronneberger2015u}
Olaf Ronneberger, Philipp Fischer, and Thomas Brox.
\newblock U-net: Convolutional networks for biomedical image segmentation.
\newblock In {\em Medical image computing and computer-assisted intervention--MICCAI 2015: 18th international conference, Munich, Germany, October 5-9, 2015, proceedings, part III 18}, pages 234--241. Springer, 2015.

\bibitem{rombach2022high}
Robin Rombach, Andreas Blattmann, Dominik Lorenz, Patrick Esser, and Bj{\"o}rn Ommer.
\newblock High-resolution image synthesis with latent diffusion models.
\newblock In {\em Proceedings of the IEEE/CVF conference on computer vision and pattern recognition}, pages 10684--10695, 2022.

\bibitem{hu2021lora}
Edward~J Hu, Yelong Shen, Phillip Wallis, Zeyuan Allen-Zhu, Yuanzhi Li, Shean Wang, Lu~Wang, and Weizhu Chen.
\newblock Lora: Low-rank adaptation of large language models.
\newblock {\em arXiv preprint arXiv:2106.09685}, 2021.

\bibitem{xFormers2022}
Benjamin Lefaudeux, Francisco Massa, Diana Liskovich, Wenhan Xiong, Vittorio Caggiano, Sean Naren, Min Xu, Jieru Hu, Marta Tintore, Susan Zhang, Patrick Labatut, Daniel Haziza, Luca Wehrstedt, Jeremy Reizenstein, and Grigory Sizov.
\newblock xformers: A modular and hackable transformer modelling library.
\newblock \url{https://github.com/facebookresearch/xformers}, 2022.

\bibitem{roich2022pivotal}
Daniel Roich, Ron Mokady, Amit~H Bermano, and Daniel Cohen-Or.
\newblock Pivotal tuning for latent-based editing of real images.
\newblock {\em ACM Transactions on Graphics (TOG)}, 42(1):1--13, 2022.

\bibitem{gal2022image}
Rinon Gal, Yuval Alaluf, Yuval Atzmon, Or~Patashnik, Amit~H Bermano, Gal Chechik, and Daniel Cohen-Or.
\newblock An image is worth one word: Personalizing text-to-image generation using textual inversion.
\newblock {\em arXiv preprint arXiv:2208.01618}, 2022.

\bibitem{https://doi.org/10.48550/arxiv.2204.11824}
Andreas Blattmann, Robin Rombach, Kaan Oktay, and Björn Ommer.
\newblock Retrieval-augmented diffusion models, 2022.

\bibitem{ho2020denoising}
Jonathan Ho, Ajay Jain, and Pieter Abbeel.
\newblock Denoising diffusion probabilistic models.
\newblock {\em Advances in neural information processing systems}, 33:6840--6851, 2020.

\bibitem{kingma2021variational}
Diederik Kingma, Tim Salimans, Ben Poole, and Jonathan Ho.
\newblock Variational diffusion models.
\newblock {\em Advances in neural information processing systems}, 34:21696--21707, 2021.

\bibitem{song2020score}
Yang Song, Jascha Sohl-Dickstein, Diederik~P Kingma, Abhishek Kumar, Stefano Ermon, and Ben Poole.
\newblock Score-based generative modeling through stochastic differential equations.
\newblock {\em arXiv preprint arXiv:2011.13456}, 2020.

\bibitem{lin2017feature}
Tsung-Yi Lin, Piotr Doll{\'a}r, Ross Girshick, Kaiming He, Bharath Hariharan, and Serge Belongie.
\newblock Feature pyramid networks for object detection.
\newblock In {\em Proceedings of the IEEE conference on computer vision and pattern recognition}, pages 2117--2125, 2017.

\bibitem{steiner2021train}
Andreas Steiner, Alexander Kolesnikov, Xiaohua Zhai, Ross Wightman, Jakob Uszkoreit, and Lucas Beyer.
\newblock How to train your vit? data, augmentation, and regularization in vision transformers.
\newblock {\em arXiv preprint arXiv:2106.10270}, 2021.

\bibitem{krizhevsky2009learning}
Alex Krizhevsky, Geoffrey Hinton, et~al.
\newblock Learning multiple layers of features from tiny images.
\newblock 2009.

\end{thebibliography}

\end{document}